# A Novel Dataset and a Deep Learning Method for Mitosis Nuclei Segmentation and Classification

Huadeng Wang[1,2], Zhipeng Liu[1], Rushi Lan[1,2], Zhenbing Liu[1,2], Xiaonan Luo[1,2], Xipeng Pan[1,2,*] and Bingbing Li[3,*]

[1]School of Computer Science and Information Security, Guilin University of Electronic Technology, Guilin, 541004, China

[2]Guangxi Key Laboratory of Image and Graphic Intelligent Processing, Guilin, 541004, China

[3]Department of Pathology，Ganzhou Municipal Hospital, Ganzhou, 341000, China

E-mail: pxp201@guet.edu.cn

**Abstract**

Mitosis nuclei count is one of the important indicators for the pathological diagnosis of breast cancer. The manual annotation needs experienced pathologists, which is very time-consuming and inefficient. With the development of deep learning methods, some models with good performance have emerged, but the generalization ability should be further strengthened. In this paper, we propose a two-stage mitosis segmentation and classification method, named SCMitosis. Firstly, the segmentation performance with a high recall rate is achieved by the proposed depthwise separable convolution residual block and channel-spatial attention gate. Then, a classification network is cascaded to further improve the detection performance of mitosis nuclei. The proposed model is verified on the ICPR 2012 dataset, and the highest F-score value of 0.8687 is obtained compared with the current state-of-the-art algorithms. In addition, the model also achieves good performance on GZMH dataset, which is prepared by our group and will be firstly released with the publication of this paper. The code will be available at: https://github.com/antifen/mitosis-nuclei-segmentation.

***Keywords: Mitosis nuclei segmentation; Mitosis nuclei classification; Depthwise separable convolution residual block; Channel-spatial attention gate; Dataset***

## 1. Introduction

The Nottingham grading system is the most widely used criterion in the grading diagnosis of breast cancer [1]. It mainly evaluates the following three aspects: (1) the degree of glandular tube formation; (2) the nuclei diversity; (3) the mitosis count. Mitosis counting is the most difficult and challenging task for several reasons: 1) Mitosis and non-mitosis nuclei have similar colors, shapes, textures, and other features, as shown in Figure 1. 2) This task is a small object segmentation task with category imbalance. The nuclei itself is a small object and compared with the non-mitosis nuclei, the number of the mitosis nuclei is scarce, presenting a quantitative imbalance between categories. 3) As the quality of pathological tissue sections is greatly affected by the production process, there often exist noise and stain contamination in pathological tissue sections. Moreover, the information contained in pathological tissue sections is complex, and the mitosis region is difficult to be completely extracted. These three reasons make the task of mitosis nuclei segmentation in breast cancer pathological images very challenging.

Most of the existing mitosis segmentation methods are designed for public datasets with a small amount of data, and they are not effective on the real-world clinical data. Therefore, the design of a mitosis region segmentation algorithm with strong generalization ability has always been a hot and difficult problem. In recent years, many methods have been proposed for the detection and counting of mitosis nuclei. Some of these methods [2, 3, 4] use feature engineering to extract the texture, shape, and other features of the nuclei for mitosis segmentation and recognition. However, due to the complex and diverse morphology of nuclei, these extracted features are often not effective in distinguishing mitosis nuclei from non-mitosis nuclei. With the wide application of convolutional neural networks in image processing tasks, many CNN-based methods have been proposed. Such as, ResNet[5] and VGG[6] are used for image classification, FCN[7], U-Net[8], and SegNet[9] are used for image segmentation, Fast-RCNN[10], Faster-RCNN[11] and Mask-RCNN[12] are used for object detection. For mitosis nuclei segmentation and classification, segmentation networks are mostly used to perform pixel-by-pixel semantic segmentation of pathological images, and thresholding and connected domain detection processing are performed on the segmented probability maps to locate mitosis regions, such as SegMitos[13] and CPCN[14].

For the task of mitosis nuclei segmentation in breast cancer pathological images, we believe that it is necessary to start from the task characteristics of small object segmentation and foreground-background category imbalance. Therefore, we propose a deep convolution mitosis nuclei segmentation model SCMitosis (Segmentation-Classification Mitosis Method). The model includes a nuclei segmentation network, SCMitosis_Seg, and a candidate sample classification network, SCMitosis_Class.

The ICPR 2012 competition dataset [15] was used to verify the performance of the model, but the amount of data is limited. In order to verify the generalization ability of the model, we also prepare a dataset named GZMH, which is from Ganzhou Municipal Hospital in Jiangxi Province, China[1]. GZMH is pixel-level annotated and double-checked by the research team's professional pathologists. It contains 1534 tiles from 22 patients, with a total of 2355 mitosis figures. This dataset is obtained from a clinical setting without additional processing. Through the verification of the GZMH dataset, we find that many existing semantic segmentation methods do not perform well, but our proposed method outperforms them.

The main contributions of this paper are as follows.

(1)We propose a novel encoder-decoder segmentation network for the segmentation of mitosis nuclei. Improved depthwise separable convolutional residual blocks (DSCRB) are used to construct the encoders and reduce the down-sampling

---

[1] The dataset is available at https://bit.ly/GZMH_Dataset.

operations to improve the segmentation ability of small targets.

(2) We propose a channel-spatial attention gate (CSAG), which is combined with a GRU unit to achieve high-quality feature fusion, reduce feature redundancy, and avoid the disappearance of deep network gradients.

(3) The proposed model is trained and tested on the ICPR 2012 dataset and achieves the best performance compared with the existing methods. In order to verify the generalization ability of the model, we also use the larger scale clinical GZMH dataset, which is released for the first time by our group, and achieve the best performance among all the comparison methods.

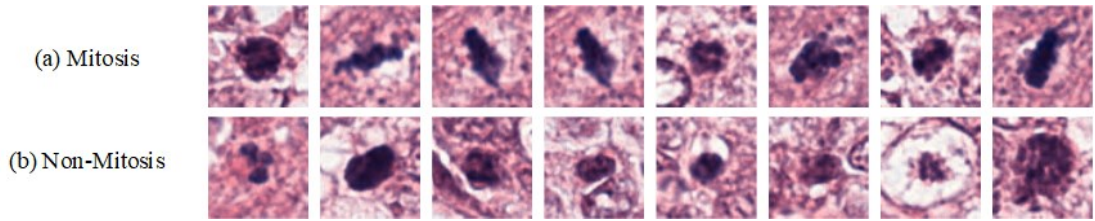

Figure 1: Examples of mitosis and non-mitosis nuclei.

## 2. Related work

As far as mitosis nuclei detection is concerned, several automated methods have been proposed [16]. In terms of the image features, we can divide them into two types, the methods based on manual feature extraction and the methods based on deep learning.

### 2.1 Methods based on the Manual Feature Extraction

Methods based on manual feature extraction usually employ staining separation, threshold detection, and fusion of geometric, morphological, and shape features [2,3,4] to detect mitosis nuclei. IPAL used a decision tree for classification after extracting features [17]. SUTECH used an object-based complete local binary pattern (CLBP) to extract texture features and used SVM for classification [18]. However, these traditional methods, which mainly rely on manually extracted features, but the expression ability of the features extracted manually is limited. Many mitosis nuclei with similar colors or special forms are difficult to be identified.

### 2.2 Methods based on the Deep Learning

#### 2.2.1 Early CNN-based Methods and Target Detection based Methods

Convolution features proved to be more powerful than manual features because they were able to automatically learn discriminative features to represent mitosis. NEC [19] designed the manual features combined with the method used by CNN and obtained an F-score of 0.659 on the ICPR 2012 mitosis dataset. On the other hand, IDSIA [20] sampled the sample points to be tested by sliding window and then used deep neural network (DNN) to predict whether it was mitosis. It obtained the highest F-score in the mitosis competition in 2012, but this detection method was not practical in clinical practice because of its large computation amount. HC+CNN [21] has designed a cascade method that requires less computational resources and makes use of hand-crafted features and convolutional features. CasNN [22] used two convolutional neural networks to form a deep cascade detection system, but it was not trained in an end-to-end way, which hindered the integration of the two networks. For example, the regional feature-based convolutional neural network (R-CNN) [23] used the proposal region generated by the selective search algorithm and used the deep convolutional feature to identify the proposal of the support vector machine, but the speed of R-CNN is very slow. In order to speed up detection, Fast R-CNN [10] was proposed, which calculated the features of the entire image and extracted the proposed feature pool layer using region of interest (RoI). However, ROIs were still generated by traditional methods, which took up a large percentage of processing time. To solve this problem, Faster R-CNN [11] was produced to generate the proposed RPN. RPN shared convolution has the function of the Fast R-CNN classification network. The DeepMitosis based on object detection [24] used the object detection network based on RPN to detect mitosis nuclei. MaskMitosis [25] used Mask R-CNN to simultaneously detect and segment mitosis nuclei.

2.2.2 Semantic Segmentation based Methods

Recently, some deep learning-based segmentation methods have been used in medical image analysis, for example, U-Net [8], which had a shortcut connection from the lower layer (contraction path) to the higher layer (expansion path) and could receive spatial information from the lower layer to make localization more accurate. The GRU recurrent neural network was considered to be a variant of LSTM [27] since only two gating structures were used, and the model complexity and computation cost of GRU were less than that of LSTM. The R2U-Net [28] combined the simplified recurrent neural network idea with U-Net, added residual blocks, and segmented skin cancer slice data, but its segmentation effect was not well at the edge of the nuclei. The U-Net++ model [29] made a deep improvement for the skip connections of U-Net, but each node needed to retain the previously fused information when it performed feature fusion, so the number of channels would be doubled. The U-Net3+ [30], was also an improved model for U-Net skip connections and fusion of multi-scale features. However, if the features of a branch were insufficient, the feature fusion of this node would be greatly affected. The $U^n$-Net [31] made more efficient use of the features extracted in the down-sampling stages, but the output of each layer in the down-sampling stages was passed backward by skip connections, which doubled the memory cost of the model, and the dilated convolution only had a significant enhancement effect on large-size objects. Considering the scarcity of medical image data, it is also a trend to use fewer samples for training. Meta multi-task learning model [32] provided a new idea for solving this task. In order to reduce the parameters and build a lightweight segmentation network, Linknet34 [33] was proposed to reduce the number of channels in the decoder feature fusion by using point-wise addition. In contrast, DeeplabV3+ [34] used dilated convolution and depthwise separable convolution to expand the receptive field without reducing the resolution of the feature maps in the later stages. SegMitos [13] used FCN [7] to construct the mitosis nuclei segmentation model but did not achieve the best performance. However, CPCN [14] used PSPNet and spatial loss function to segment mitosis nuclei and achieved good performance.

With the development of semantic segmentation networks, great progress has been made in such problems as multi-scale feature extraction and large target segmentation. However, there are still some defects in feature fusion, small target segmentation and so on, which are the main problems that we pay attention to in our current work.

## 3. Method

Figure 2. (a) shows our SCMitosis model, which includes two parts. A mitosis nuclei segmentation network, SCMitosis_Seg, is used to segment mitosis nuclei regions and extract mitosis nuclei candidate samples, as shown in Figure 2. (b). A candidate sample classification network, SCMitosis_Class, is used for further refined classification of candidate samples extracted by SCMitosis_Seg, as shown in Figure 2. (c).

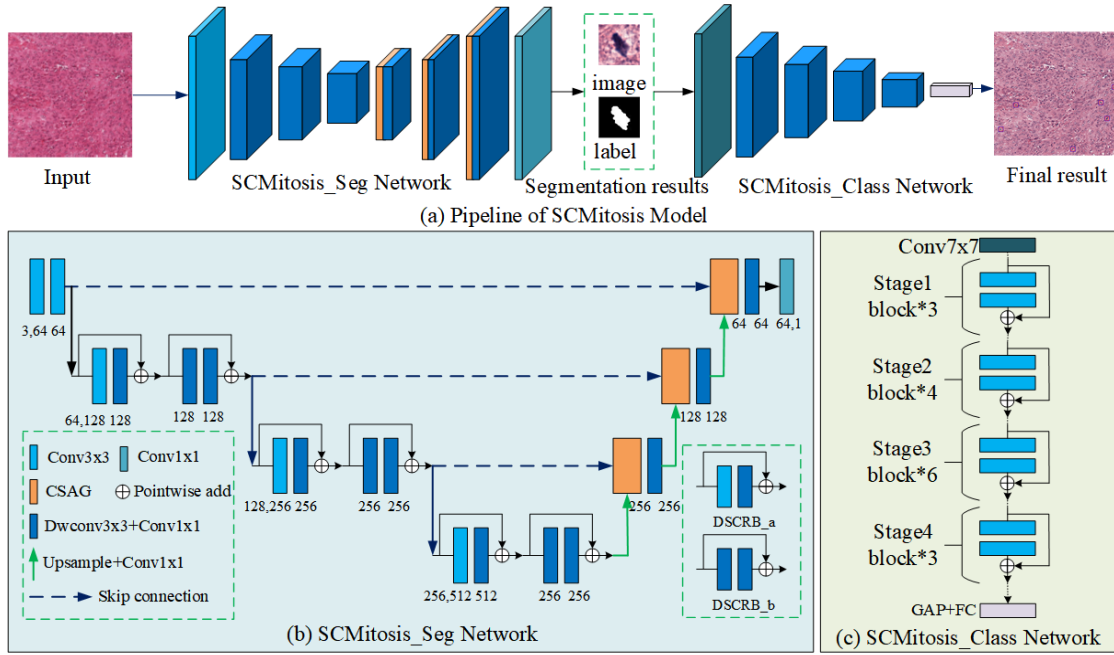

Figure 2: Overview of SCMitosis Model. (a) The pipeline of SCMitosis model architecture. SCMitosis first uses the mitosis nuclei segmentation model SCMitosis_Seg. The candidate mitosis nuclei regions were segmented and then input into the mitosis nuclei classification model SCMitosis_Class to refine the segmentation results. (b) SCMitosis_Seg network diagram. In this network, DSCRB is used in an encoder, CASG is used in a decoder for feature fusion, and the network reduces the downsampling times to improve the performance of small target segmentation. (c) SCMitosis_Class network diagram. In this network, the residual blocks are combined and the global average pooling operation is used.

## 3.1 Mitosis Segmentation Network

### 3.1.1 Depthwise Separable Convolution Residual Block

SCMitosis_Seg is a fully convolutional semantic segmentation network, which is divided into encoder and decoder phases. As shown in Figure 2. (b), in order to enhance the feature extraction ability, we use residual blocks to form the encoders. Although the increase in the number of convolutional layers and the number of skip connections can improve the ability of model feature extraction, the increase is not linear, and it has the same boundary-decreasing effect as the increase of training data. We only started using the residual structure in the second encoder and still used two ordinary 3x3 convolutions instead of residual blocks in the first encoder, and we also used depthwise separable convolution [35] to reconstruct the residual block, which was named depthwise separable convolution residual block (DSCRB). By using DSCRB, the number of parameters in the replaced convolutional layer can be reduced to about 1/8 of the original, which effectively reduces the complexity of the model.

As shown in Figure 3, the proposed depthwise separable convolution residual blocks are divided into two types, DSCRB_a and DSCRB_b. The main difference between the two is whether the first convolution layer of each residual block is down-sampled. If the first convolution layer of the residual block is down-sampled, this layer is still the ordinary 3x3 convolution layer, and the second convolution layer is replaced by depthwise separable convolution, while the 1x1 convolution used to adjust the number of channels during the addition operation remains unchanged. This depthwise separable convolution residual block is called DSCRB_a. If the first convolution layer of the residual block is not down-sampled, then the residual block contains only two convolution layers and both will be replaced by depthwise separable convolution. This depthwise separable convolution residual block is called DSCRB_b.

When DSCRB is used to construct an encoder, a DSCRB_a and a DSCRB_b are connected in series to form a depthwise separable convolution residual encoder (DSCR_Encoder). Due to every time after down-sampling, the size of the feature map is reduced to 1/4 of the original feature map, considering the small size of the mitosis nuclei, we think that this is a small object segmentation task, and if too many down-sampling operations are performed, the mitosis nuclei may not occupy a single pixel in the final feature map. Therefore, we decided to perform only three down-sampling operations in the encoder part. Because the first encoder is only composed of ordinary 3x3 convolution, three DSCRB_Encoders are included in the encoder part.

In the decoder part, up-sampling is also carried out three times, and considering the "checkerboard effect" caused by the use of transposed convolution [36], that is, the use of transposed convolution will produce an obvious superposition phenomenon at the position of convolution kernel superposition. Because breast cancer mitosis nuclei segmentation is a small object task, it is sensitive to the checkerboard effect. Therefore, instead of using transpose convolution, bilinear interpolation is employed to perform upsampling, and a $1 \times 1$ convolution is added to adjust the number of channels. In order to better carry out feature fusion, we also propose a channel-spatial attention gate (CSAG) in the decoder part, which will be introduced in Section 3.1.2. In addition, the subsequent convolutional layer of the decoder only uses depthwise separable convolution (DSC) instead of DSCRB, which is mainly to reduce the number of parameters and model complexity.

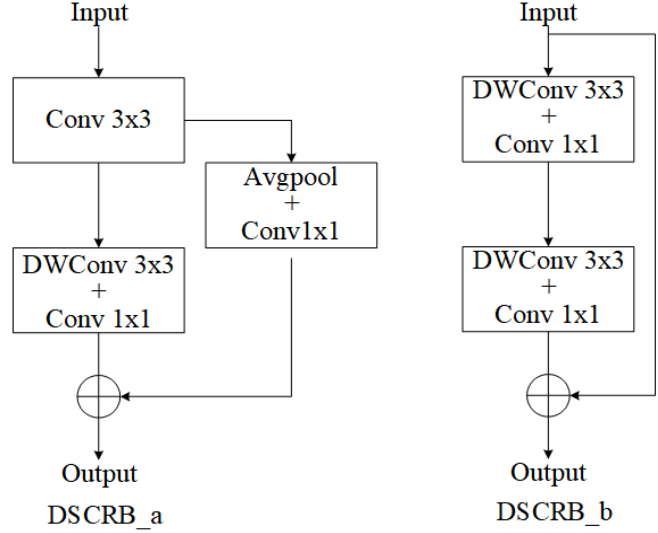

Figure 3. Example of Depthwise Separable Convolution Residual Block (DSCRB).

### 3.1.2 Feature Fusion Based on Attention Gate

To better strengthen the features, we use channel-spatial feature attention to focus on both channel and spatial features and focus on the features extracted by each encoder. In the feature fusion part, compared with the simple channel direct addition or channel concatenation, and referring to the solution of the NLP (natural language processing) task, we employ the GRU gating unit with a relatively simple structure [25] to form the feature fusion module in the decoder part of the segmentation network. This is the first time GRU has been applied to mitosis nuclei segmentation.

Different from the direct use of the Convolutional Block Attention Module (CBAM) [37] and GRU unit cascade, we propose a Channel-Spatial Attention Gate (CSAG) (Figure 2. (c)). Specifically, we first extract channel attention features CEi and CDi from the output of each encoder Ei and each decoder Di, and then input these two features into GRU for feature fusion and update to obtain the fusion channel attention features CFi. Similarly, spatial attention SEi and SDi are extracted from Ei and Di, and input to GRU to obtain the integrated spatial attention features SFi. Then, Ei and Di are added and then multiplied with CFi and SFi to complete feature enhancement and feature fusion. For details, see Eq. (1) to Eq. (5).

$$CEi, CDi = Channel\_Attention(Ei, Di) \qquad (1)$$

$$CFi = GRU(CEi, CDi) \qquad (2)$$

$$SEi, SDi = Spatial\_Attention(Ei, Di) \qquad (3)$$

$$SFi = GRU(SEi, SDi) \qquad (4)$$

$$CSFi = (Ei + Di) \times CFi \times SFi \qquad (5)$$

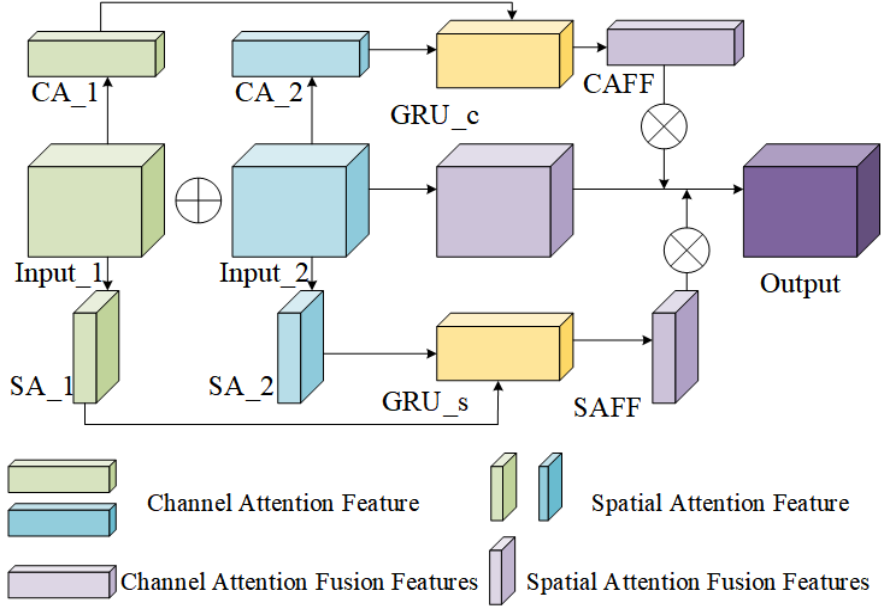

Figure 4. Channel-Spatial Attention Gate (CSAG). The output of the corresponding encoder and the previous decoder are fused with the features of the channel and spatial attention.

### 3.1.3 Loss and Evaluation

Our evaluation metrics mainly include recall, precision and F-score. However, according to Eq. (8), F-score is affected by both precision and recall, and in reality, models tend to fail to obtain very high values of precision and recall at the same time. Therefore, recall value in the segmentation stage should be considered more. Because there is also a classification loss at the classification stage, the segmentation model should segment as many mitosis nuclei as possible. Considering that this is a small object segmentation task with imbalanced categories, we use the combined loss function of BCE loss and Tversky loss as in Eq. (9), which can maintain a higher recall value, and the precision will not become extremely low. After comparison, we choose the coefficient of 0.3 and 0.7 to control the weight of Bce_loss and Tversky_loss, and the values of α and β in Tversky_loss are 0.3 and 0.7, respectively.

$$recall = TP/(TP + FN) \qquad (6)$$

$$precision = TP/(TP + FP) \qquad (7)$$

$$F - score = \frac{2 \times precision \times recall}{precision + recall} \qquad (8)$$

$$Loss = 0.3 \times Bce\_loss + 0.7 \times Tversky\_loss \qquad (9)$$

### 3.1.4 Candidate Sample Extraction

We first use a 2048x2048 window to slide and crop the 2084x2084 image four times (2x2) as the prediction input and take the average value of the prediction in the overlapping area by overlapping times. Then, after detecting the connected domain of the binarized segmentation probability map, the segmented nuclei centroid and area are calculated, and the nuclei with an area greater than 100 pixels is cropped according to the center as the candidate sample.

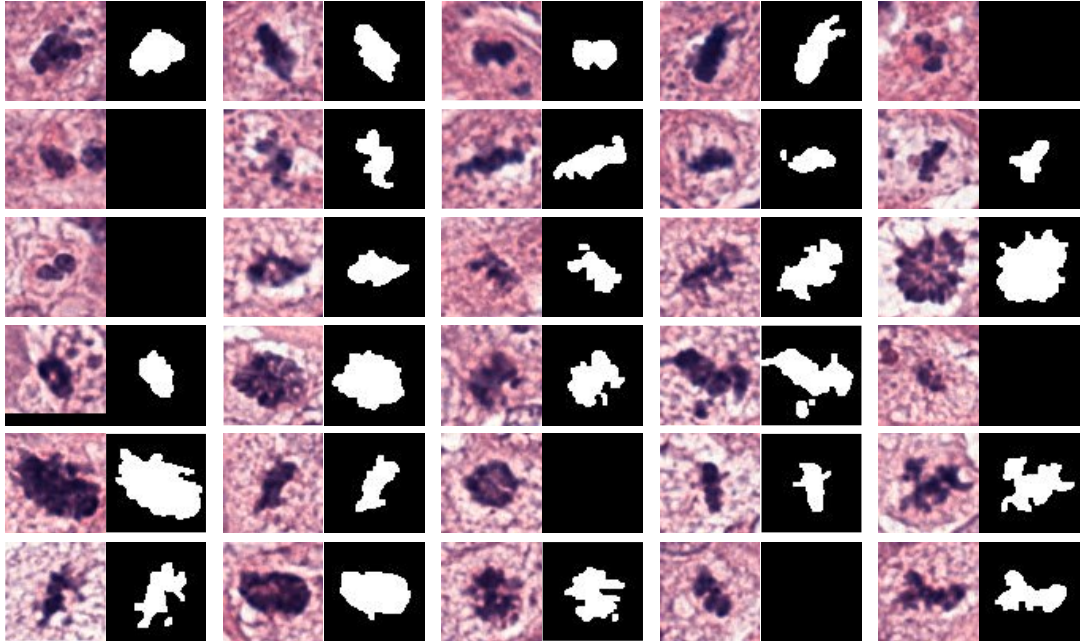

Figure 5. Candidate samples and their corresponding labels. The left side of each column shows the candidate samples, and the right side shows the corresponding labels.

### 3.2 Candidate Sample Classification Network

In this stage, we select several classification models such as ResNet18/34/50 [5], MobileNetV3_Large [39], and Densenet201 [40] which are pre-trained on ImageNet dataset [38] to perform classification experiments on candidate samples. In terms of specific details, we enlarge the size of candidate samples to 128x128 and then input them into the network for training and prediction. This is mainly to reduce the influence caused by excessive downsampling times in the classification network. In addition, there are many online image enhancement methods in the pre-processing stage. Since SCMitosis_Seg has finely segmented most of the targets, most of the negative samples are filtered out, which effectively reduces the adverse impact caused by the imbalance of positive and negative samples faced by the classification model. As shown in Figure 6, we select ResNet34 as our SCMitosis_Class model.

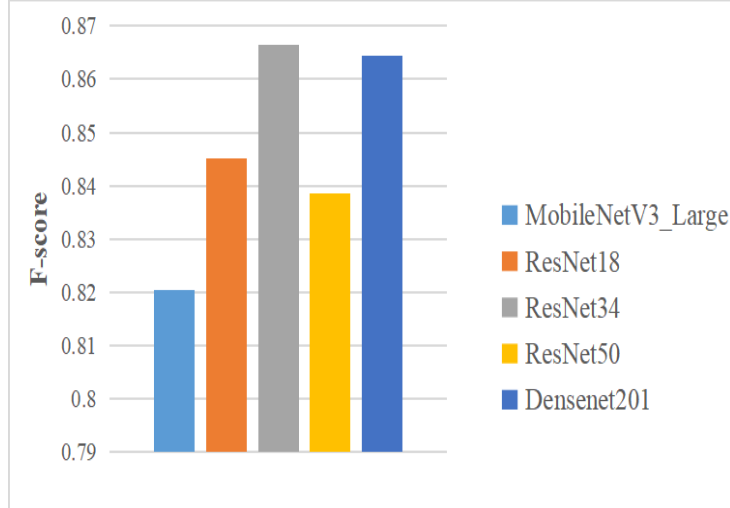

Figure 6. Performance of different classification models on the ICPR 2012 dataset.

## 4. Experiment

### 4.1 Experimental Dataset

4.1.1 ICPR 2012 Dataset

The 2012 ICPR MITOSIS dataset [15] uses five H&E-stained breast cancer biopsy sections. Each section is scanned using an Aperio ScanScope XT slide scanner, generating 10 HPF per section at 40X magnification for a total of 50 HPFs. The size of each HPF is 2084×2084 pixels, which corresponds to a microscope area of 512μm×512μm. Experienced pathologists manually labeled all mitosis pixels of these 50 HPF. A total of 35 HPFs with 206 annotated mitosis constitute the training set, and the remaining 15 HPFs with 101 mitosis nuclei are used for testing.

4.1.2 GZMH Dataset

This dataset is first released by our group. The image data in this dataset are all from Ganzhou Municipal Hospital, and the outlines of the mitosis images are finely annotated by professional pathologists, and they are double-checked. All data are scanned using a digital section scanner (KF-PRO120) at 40x magnification with a resolution of 0.25um/pixel. WSI is converted to SVS format, and then the component features of the mitosis image of tumor cells are manually labeled. The mitosis data of breast cancer are manually labeled using the multi-stage physician review system to determine the mitosis data. For example, the mitosis data are labeled by 3 pathologists (1 resident, and 2 attending doctors; All of them have more than 5 years of working experience), and are reviewed by 2 senior pathologists (2 chief doctors; More than 10 years of working experience); The final labeling results are reviewed by 2 senior pathologists. If the labeling results are different, 5 pathologists will reprint the film to determine whether it is a mitosis image. The dataset contains 55 WSIs from 22 patients. It is divided into two parts: the training set and the test set. The training set contains 48 WSIs from 20 patients, and the test set contains 7 WSIs from 2 patients.

The specific form is 1534 RGB channel electronic images with a resolution of 2084x2084 pixels, and their corresponding single-channel black and white binary labels. The training set contained 1192 HPF images with a total of 1832 mitosis regions. The test set consisted of 342 HPF images with a total of 523 mitosis regions, and the training and test sets are from different patients without crossover. The specific information of the GZMH dataset and several classical mitosis nuclei detection datasets is shown in Table 1 and Figure 7 shows the images of the GZMH dataset. We are the first team to use this dataset.

Table 1. Comparison of several typical mitosis nuclei detection datasets with the GZMH dataset. "NA" indicates that the test set of the MITOS-ATYPIA-14 dataset contains an undisclosed number of mitosis nuclei.

| Dataset | Mitosis numbers | WSI images | HPF images | HPF size |
|---|---|---|---|---|
| 2012- MITOS[15] | 327 | 5 | 50 | 2084 x 2084 pixel |
| AMIDA13[43] | 1083 | 23 | 606 | 2000 x 2000 pixel |
| MITOS-ATYPIA-14[42] | 749+NA | 21 | 1696 | 1539 x 1376 pixel |
| GZMH | 2355 | 55 | 1534 | 2084 x 2084 pixel |

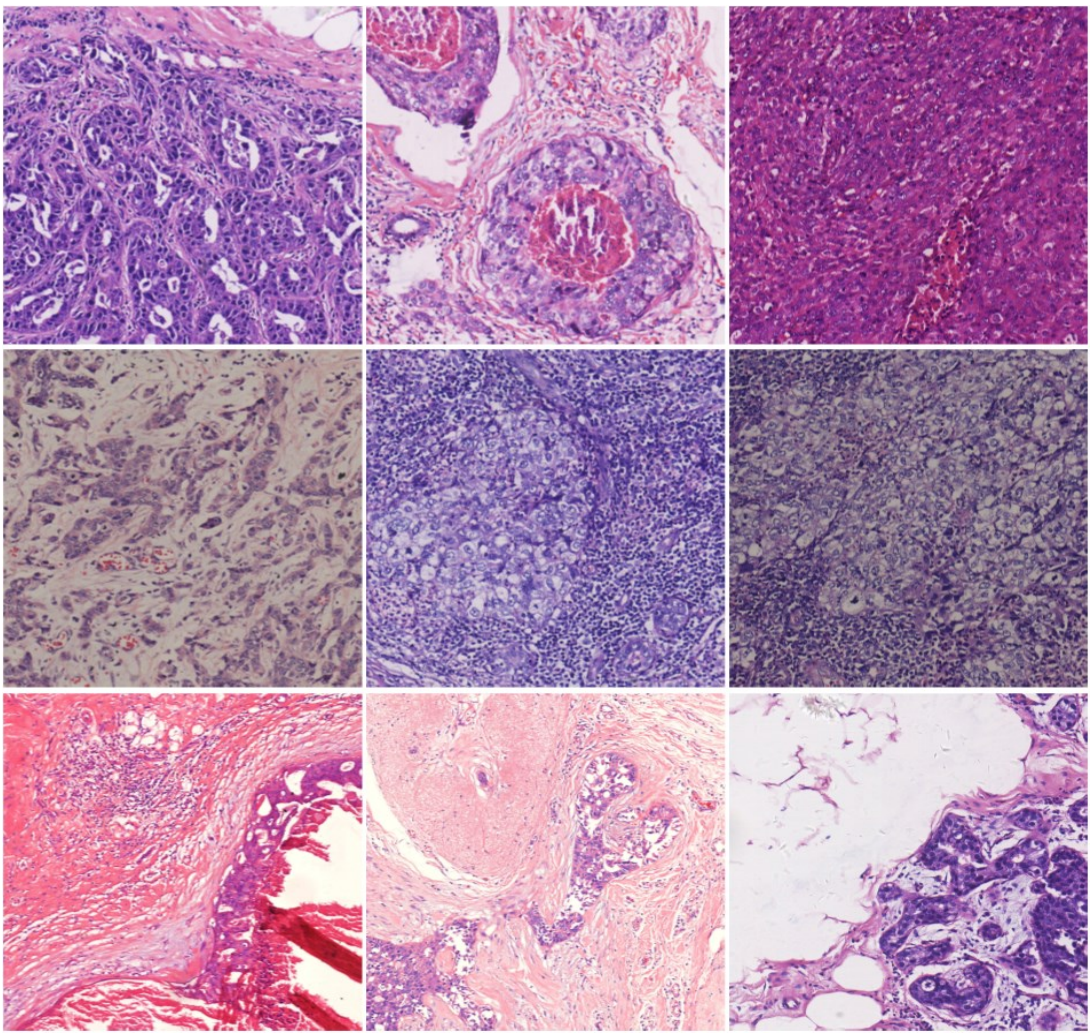

Figure 7. Sample images from the GZMH dataset.

4.1.3 Data Pre-processing and Augmentation

We perform the same pre-processing operations on both datasets, as shown in Figure 8. first, all images are performed by staining normalization operation [43] to reduce the adverse effects caused by color differences. Then, the images are cropped into patches with the size of 256x256, which can be divided into sliding window cropping with 32-pixel overlap, random cropping, and cropping around the center coordinates of the mitosis image. Finally, online random flip, rotation, filtering, re-scaling and cropping, and other data enhancement methods are also used.

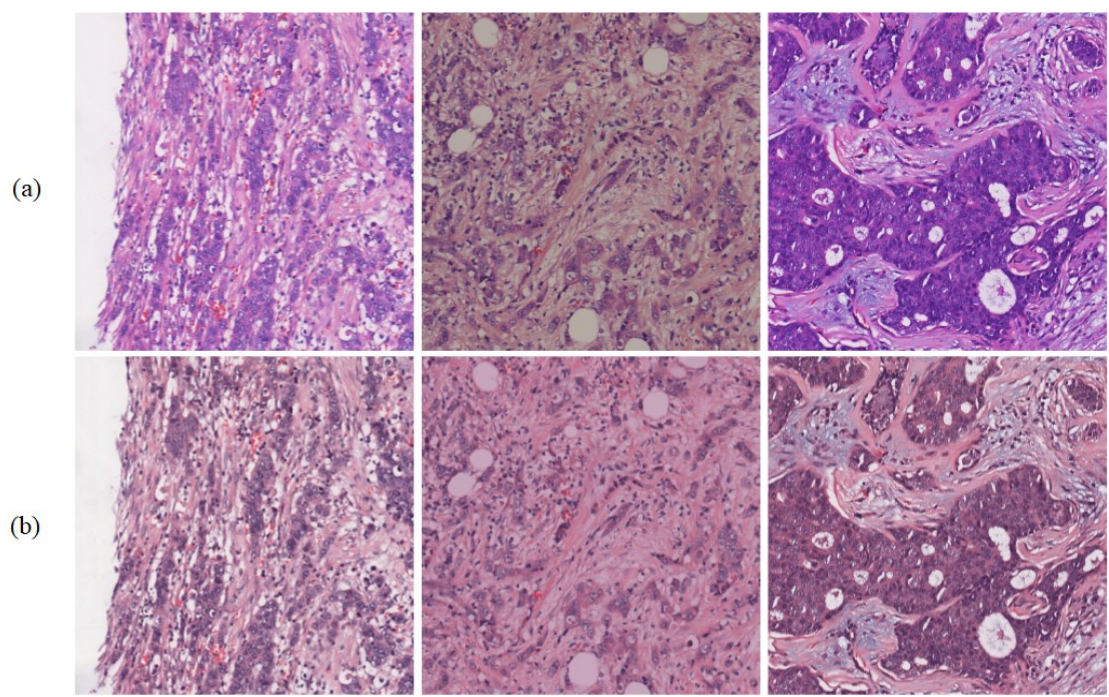

Figure 8. An example of staining normalization operation used for images in the GZMH dataset. (a) the original images, and (b) the images after staining normalization.

4.1.4 Experimental Setup

In the training process, we use the AdamW optimizer with an initial learning rate of 1e-4, a decay ratio of 0.1, and a batch size of 16. The training is carried out on a Telsa V100 32G GPU.

4.2 Ablation Experiments

We perform ablation experiments on the ICPR 2012 dataset by adding different modules to SCMitosis_Seg to study the effect on the performance improvement of the model. As shown in Table 2, based on the method of using depthwise separable convolution (DSC) and CBAM , we can see that the recall rate is high at this time, but the precision is low, which means that a large number of false positive samples are segmented. The number of false positive samples can be effectively reduced when the depthwise separable convolution component is replaced by depthwise separable convolution residual block (DSCRB) or GRU gating unit is added to the basic model, but it also leads to certainly missed detection. In this case, the F-score increased significantly, indicating that the residual structure extracted more effective features and GRU unit can carry out feature fusion with higher quality. However, when we use both residual depthwise separable convolution block (RSC) and GRU gated unit, recall is further improved, but precision is decreased, which may indicate that component stacking cannot bring continuous performance improvement. However, when the proposed channel-spatial attention gate is used to replace the CBAM-GRU cascade structure, the segmentation performance of the model is significantly improved and the highest F-score is obtained, which indicates that the proposed CSAG attention gating unit is more effective than the simple stacked CBAM-GRU cascade structure.

Further, as shown in Table 1, candidate samples obtained from the segmentation model with the best performance on the ICPR 2012 dataset are used to compare several classification models. ResNet34 obtains the highest F-score with fewer parameters that compared with ResNet50, and ResNet34. This also shows that simply stacking and augmenting models does not necessarily lead to improved performance.

Table 2. First-stage segmentation results of the SCMitosis_Seg using different modules on the ICPR 2012 dataset.

| Module usage | | | | | Evaluation metrics | | |
|---|---|---|---|---|---|---|---|
| DSC | CBAM | GRU | DSCRB | CSAG | Precision | Recall | F1-score |
| √ | √ | | | | 0.2069 | 0.9505 | 0.3398 |
| √ | √ | √ | | | 0.3973 | 0.8812 | 0.5477 |
| | √ | | √ | | 0.4387 | 0.9208 | 0.5942 |
| | √ | √ | √ | | 0.2939 | **0.9604** | 0.4501 |
| | | | √ | √ | **0.5137** | 0.9307 | **0.6620** |

## 4.3 Performance on the ICPR 2012 dataset

SCMitosis achieves the best performance on the ICPR 2012 dataset. Table 3 shows the details of the proposed method compared with other methods. Among them, IDSIA [20], IPAL [17], SUTECH [18], and NEC [19] are among the top four methods in ICPR 2012 mitosis Detection Competition. The other methods in Table 3 are proposed after the ICPR 2012 Mitosis Detection Competition. Our method achieves a precision value of 0.8766, a recall value of 0.8515, and the highest F-score of 0.8643 in the ICPR 2012 dataset, which indicates that our method is the state-of-the-art. Figure 9 shows the final segmentation detection result. Figure 10 shows the false positive candidate samples correctly identified by SCMitosis_Class, which indicates that our proposed SCMitosis can be applied to the field of mitosis detection.

Table 3. The experimental results of different methods on the ICPR 2012 dataset.

| Model | Precision | Recall | F1-score |
|---|---|---|---|
| NEC (Malon et al., 2013)[19] | 0.750 | 0.590 | 0.6592 |
| SUTECH (Tashk et al., 2013) [18] | 0.700 | 0.720 | 0.7094 |
| IPAL (Irshad et al., 2013) [17] | 0.6981 | 0.740 | 0.7184 |
| IDSIA (Cire¸san et al., 2013) [20] | 0.886 | 0.700 | 0.782 |
| HC+CNN (Wang et al., 2014)[21] | 0.840 | 0.650 | 0.7345 |
| CasNN (Chen et al., 2016a) [22] | 0.804 | 0.772 | 0.788 |
| RRF (Paul et al., 2015) [41] | 0.835 | 0.811 | 0.823 |
| DeepMitosis (Li et al., 2018) [24] | 0.854 | 0.812 | 0.832 |
| SegMitos-8s (Chao et al., 2019)[13] | 0.8461 | 0.7624 | 0.8021 |
| MaskMitosis (Meriem et al., 2020)[25] | **0.921** | 0.811 | 0.863 |
| CPCN(Han et al., 2021)[14] | 0.8447 | **0.8614** | 0.8529 |
| SCMitosis(ours) | 0.8776 | 0.8515 | **0.8643** |

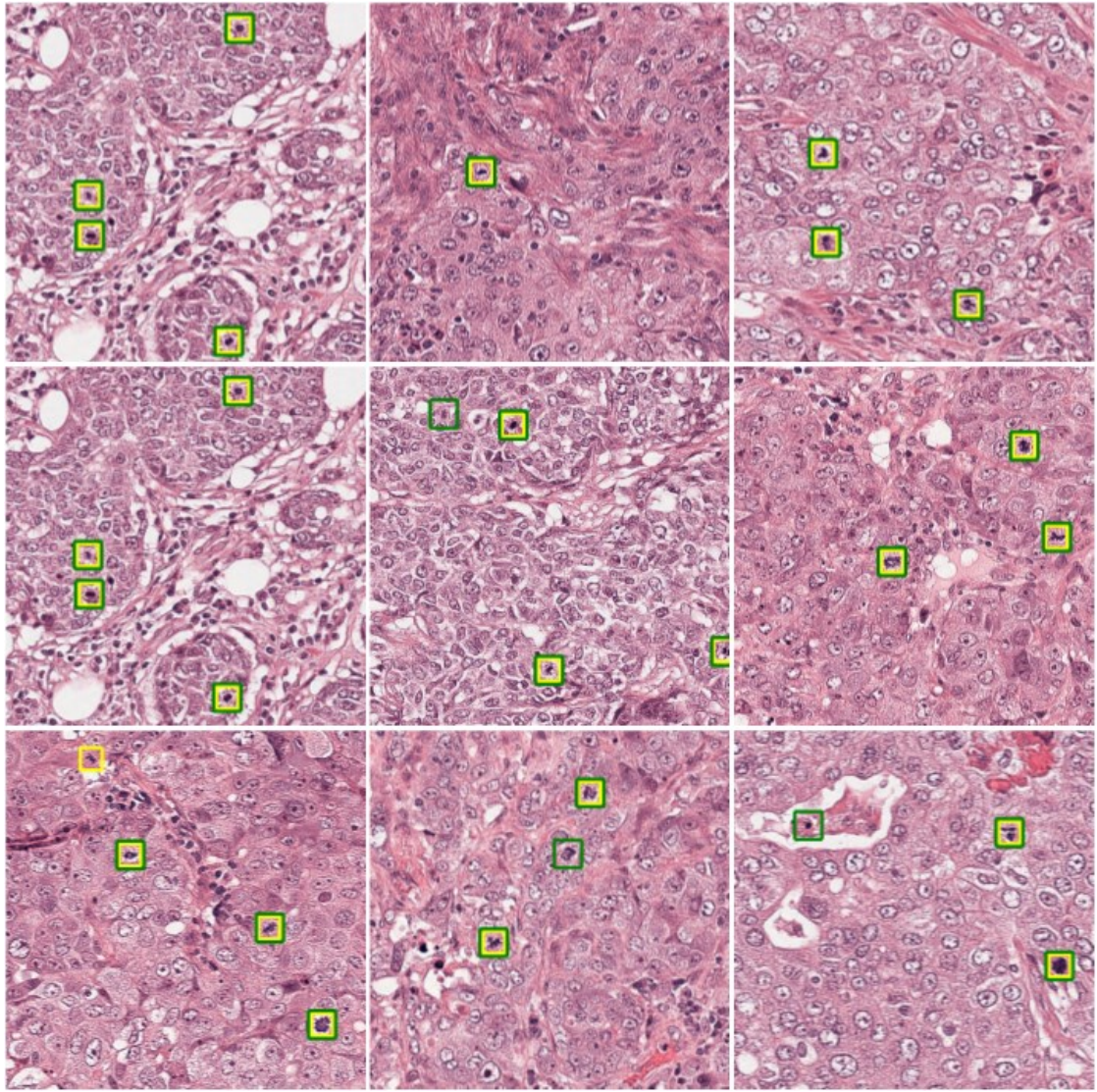

Figure 9. Detection results of ICPR 2012 dataset. The green box shows the mitosis nuclei as believed by the segmentation network, and the yellow box shows the location results marked by pathologists. It can be seen that the distribution of mitosis nuclei in pathological images is very sparse, a large number of areas are background areas, and most of the nuclei are also non-mitosis.

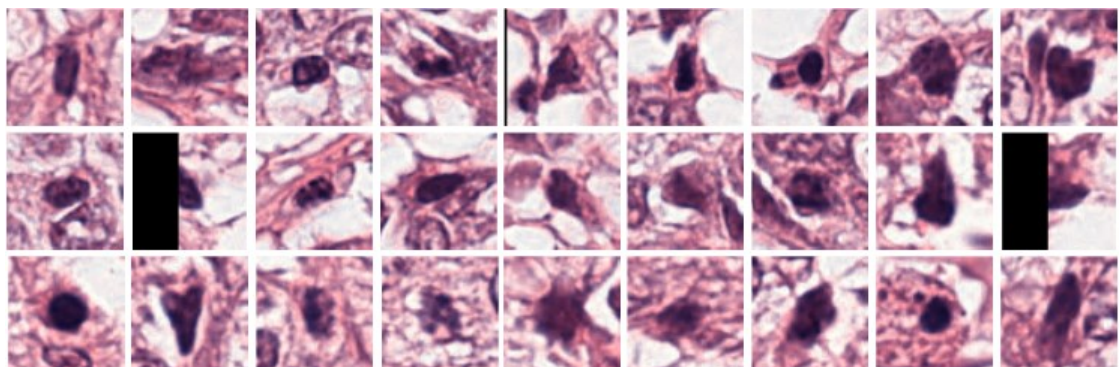

Figure 10. Examples of false positive candidate samples identified by the classification model. The black area in the image is the false positive candidate samples located in the boundary area of HPF.

## 4.4 Performance on the GZMH dataset

Furthermore, we use the model with the best performance on the ICPR 2012 dataset to train and validate on the GZMH dataset. Since it is the first application of this dataset, we select several segmentation methods for the same semantic segmentation task to participate in the comparison. As shown in Table 4, the results of our proposed method also obtain the highest F-score among the compared methods, which also proves the effectiveness of the proposed method in this paper. Figure 11 shows the final detection results of the GZMH dataset. The performance of other methods on this dataset also indicates that the task of breast cancer mitosis nuclei segmentation on larger datasets is very challenging.

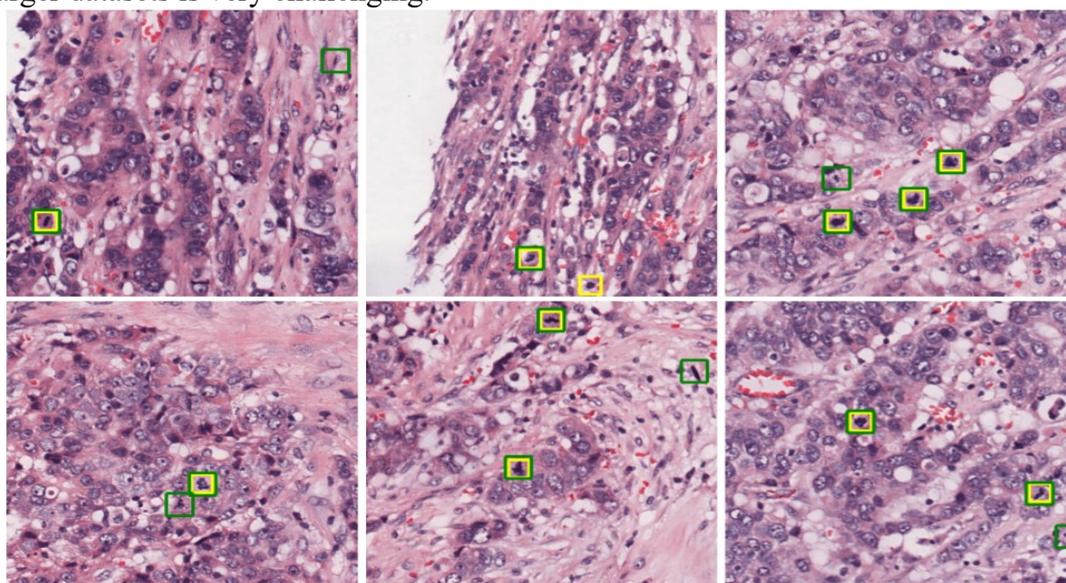

Figure 11. Display of the GZMH dataset detection results. The green box shows the mitosis nuclei as believed by the segmentation network, and the yellow box shows the location results marked by pathologists.

Table 4. Experiment results of different methods on the GZMH dataset

| Model | Precision | Recall | F1-score |
|---|---|---|---|
| U-Net[8] | 0.3022 | 0.7784 | 0.4353 |
| SegNet[9] | 0.2904 | **0.8304** | 0.4304 |
| R2U-Net[28] | 0.3164 | 0.8054 | 0.4543 |
| LinkNet34[33] | 0.3623 | 0.7225 | 0.4826 |
| DeepLabV3+[34] | 0.2937 | 0.7900 | 0.4282 |
| SCMitosis(ours) | **0.4278** | 0.7325 | **0.5402** |

## 4.5 Discussion

Although our method has achieved the highest F-score on both ICPR 2012 dataset and GZMH dataset, the performance on the two datasets is not consistent. On the ICPR 2012 dataset, the F-score of our method achieves 0.8687, while on the larger GZMH dataset, the F-score is only 0.5402, and on the GZMH dataset, the scores of the other methods in the comparison are even lower. Due to the total number of mitosis nuclei contained in the GZMH dataset, the number of cases and the total number of HPF are larger, and the distribution of mitosis nuclei is more sparse and more complex. Considering this manifestation of data differences, we believe that the segmentation of mitosis nuclei in the larger-scale breast cancer pathology image data represented by the GZMH dataset is an extremely challenging task.

## 5. Conclusion

In this paper, we propose a method for the segmentation and classification of mitosis nuclei in HE-stained breast cancer pathological images. Our approach is based on a two-stage strategy combining encoding and decoding structure semantic segmentation and classification and achieves the optimal performance on the pixel-level annotated public ICPR2012 dataset. We also release a pixel-level annotated GZMH dataset by professional pathologists. Our method achieves the best performance on the GZMH dataset, which indicates that our method has good generalization performance and clinical application potential. In the future, we hope to establish a complete set of mitosis nuclei segmentation, classification, and automated analysis methods to better assist doctors in breast cancer diagnosis.